\documentclass[12pt]{article}
\usepackage{amsfonts}

\newtheorem{theorem}{Theorem}[section]
\newtheorem{lemma}[theorem]{Lemma}
\newtheorem{corollary}[theorem]{Corollary}
\newtheorem{rmk}[theorem]{Remark}
\newtheorem{dfn}[theorem]{Definition}

\newcommand{\lppt}{{\ll}}
\newcommand{\rppt}{{\gg}}
\newcommand{\oH}{\overline{H}}

\newcommand{\calB}{\mathcal{B}}
\newcommand{\calC}{\mathcal{C}}
\newcommand{\calI}{\mathcal{I}}
\newcommand{\calF}{\mathcal{F}}

\newcommand{\calL}{\mathcal{L}}

\newcommand{\calS}{\mathcal{S}}
\newcommand{\bbR}{\mathbb{R}}

\newcommand{\Adual}{A^{*}}
\def\Im{\mathop{\rm Im}\nolimits}
\newcommand{\ract}{\mathbin{{\leftharpoonup}}}
\newcommand{\lact}{\mathbin{{\rightharpoonup}}}
\newcommand{\Ker}{\mathop{\rm Ker}}
\newcommand{\jperp}{J^{\perp}}
\newcommand{\ebar}{{\bar e}}

\newcommand{\xbar}{{\bar x}}
\newcommand{\w}[1]{{\mathcal{W}(#1)}}
\newcommand{\pf}{\medskip\noindent{\sc Proof: }}

\title{State Space Realization Theorems \protect\\ For Data Mining}

\author{Robert L. Grossman\thanks{Corresponding author:
University of Illinois at Chicago,
Department of Mathematics, Statistics \& Computer Science (MC 249),
851 S. Morgan Street, Chicago IL 60607, grossman@uic.edu,
312 413 2176} \ and Richard G. Larson \protect\\ 
University of Illinois at Chicago}

\date{December 19, 2008}

\begin{document}

\maketitle

\vfill\eject

\begin{abstract}
In this paper, we consider formal series associated with events,
profiles derived from events, and statistical models that make
predictions about events.  We prove theorems about realizations for
these formal series using the language and tools of Hopf algebras.

\medbreak
\end{abstract}

\vfill\eject

\section{Introduction}

Many data mining problems can be formulated in terms of events,
profiles, models and predictions.  As an example, consider the problem
of predicting credit card fraud.  In this application, there is a
sequence of credit card transactions (called the learning set), each
of which is associated with a credit card account and some of which
have been labeled as fraudulent.  The goal is to use the learning set
to build a statistical model that predicts the likelihood that a
credit card transaction is associated with a fraudulent account.
Information about each credit card transaction is aggregated to
produce a statistical profile (or state vector) about each credit card
account.  The profile consists of features. Applying the model to the
profile produces a prediction about whether the account is likely to
be fraudulent.  Note that we can think of this example as a map from
inputs (events) to outputs (predictions about whether the associated
account is fraudulent).  Given such an input-output map, we can ask
whether there is a ``realization'' in which there is a state space of
profiles (corresponding to accounts) in which each event updates the
corresponding profile.  We will see how to make this precise below.

Usually several different fraud models are developed and compared to
one and another.  Each fraud model is associated with a
misclassification rate, which is the percent of fraudulent accounts
that remain undetected.  For many data mining applications, especially
large-scale applications, we do not have a single learning set, but
rather a collection of learning sets.

In this paper, we abstract this problem and use the language and tools
of Hopf algebras to study it.  To continue the example above, we
abstract credit card transactions as {\em events}; state information
about credit card accounts as {\em profiles}; credit card account
numbers as {\em profile IDs or PIDs}; statistical models predicting
the likelihood that a credit card account is fraudulent as {\em
  models}; a sequence of credit card transactions each of which is
labeled either valid or fraudulent as {\em learning sets} of labeled
events; and the accuracy rate of the credit card fraud model as the
{\em classification rate} of the model.

We are interested in the following set up.  Consider a collection
${\cal C}$, possibly infinite, of labeled learning sets $w$ of events.
For a labeled learning set $w$, we can build a model.  Each model has
a classification rate $p_w$.  This information can be summarized in a
formal series
$$ p = \sum_{\cal C} p_w w$$
In this paper, we prove some theorems about these formal series
using the language of Hopf algebras.

We now give the precise definitions we need.  A {\em labeled event} is
an event, together with a Profile Identifier (PID) and a label.
Fix a set $D$ of labeled events.
We define a {\em labeled learning set of events\/} to be an element of $\w{D}$,
the set of words $d_1\cdots d_k$ of elements $d_i\in D$.
If $k$ is a field, then $H=k\w{D}$ is a $k$-algebra with basis $\w{D}$.
In this paper we study formal series of the form
\[
 p = \sum_{w\in \w{D}} p_w w.
\]
By a formal series, we mean a map
\[
H \longrightarrow k,
\]
associating to each element $w\in \w{D}$ the series coefficient
$p_w$.
The coefficient $p_w$ is the classification (or misclassification) rate
for the learning set of the events in $w$.
Formal series occur in the formal theory of languages, automata theory,
control theory, and a variety of other areas.

There is a more concrete realization of a model that
we now describe.  This requires a space $X$ whose points
$x\in X$ we interpret as
profiles or states, which abstract the features used
in a model.  We can now define a model
as a function from 
a space $X$ of profiles that assigns a label (in $k$) 
to each element $x\in X$:
\[
f:X\longrightarrow k.
\]
Notice that given an initial profile $x_0\in X$ associated
with a PID, a sequence of
events associated with a single PID will sweep out an orbit 
in $X$ since each event will update the current profile in $X$
associated with the PID.  In the paper, we usually call
the space $X$ the {\em state space} and the initial profile
the {\em initial state}.  

Fix a formal series $p$. We investigate a standard question: given
a formal series $p$ built from the events $D$, is there
a state space $X$, a (classification) model
\[
f:X\longrightarrow k,
\]
and a set of initial states that yield $p$.  This is called
a realization theorem.
The state space captures the ``essential'' information in the data which
is implicit in the series $p$.
The formal definition is given below.

Realization theorems use a finiteness condition to imply the infinite
object can be represented by a finite state space.  One of the most
familiar realization theorems is the Myhill--Nerode theorem.  In this
case, the infinite object is a formal series of words forming a
language; the finiteness condition is the finiteness of a right
invariant equivalence relation, and the state space is a finite
automaton.  In the case of data mining, the infinite object is a
formal series of learning sets comprising a series of experiments, the
finiteness condition is described by the finite dimensionality of a
span of vectors, and the state space is ${\bbR}^{n}$.

The Myhill--Nerode theorem and more generally languages, formal
series, automata, and finiteness conditions play a fundamental role in
computer science.  Our goal is to introduce analogous structures into
data mining.

We now briefly recall the
Myhill--Nerode theorem following \cite[page 65]{Hopcroft}.
Let the set $D$ be an alphabet,
$\w{D}$ be the set of words in $D$, and $L \subset
\w{D}$ be a language.  A language $L$ defines an equivalence relation
$\sim$ as follows: for $u, v\in \w{D}$, $u \sim v$ if and only if for
all $w\in \w{D}$ either both or neither of $uw$ and $vw$ are in $L$.  An
equivalence relation $\sim$ is called {\em right invariant\/} with respect to
concatenation in case $u\sim v$ implies $uw \sim vw$ for all $w\in
\w{D}$.

\begin{theorem}[Myhill--Nerode]
The following are equivalent:
\begin{enumerate}

\item $L$ is the union of a finite number of equivalence
classes generated by a right invariant equivalence relation.

\item The language $L\subset \w{D}$ is accepted by a finite automaton.

\end{enumerate}
\end{theorem}

We point out that in this case
a language $L\subset \w{D}$ naturally defines a formal series.
Fix a field $k$ and
the $k$-algebra $H=k\w{D}$. Given a language
$L$, define the formal series $p$
as follows:
\[
 p(w) = \left\{ \begin{array}{ll}
        1 & \mbox{if $w\in L$ } \\
        0 & \mbox{otherwise}
        \end{array}
        \right..
\]

Section~2 contains preliminary material.
Section~3 constructs a finite state space $X$ for the simple case of a
formal series without profile identifiers or labels.
Section~4 proves a theorem about parametrized classifiers and near to
best realizations.
Section~5 contains our main realization theorem.

One of the goals of this paper is to provide an algebraic foundation
for some of the formal aspects of data mining.  Other (non-algebraic)
approaches can be found in~\cite{Valiant}, \cite{Vapnik} and
\cite{Cucker}.

A short annoucement of the some of the results in this paper (without
proofs) appeared in \cite{GL:2002}.

\section{Preliminaries}\label{intro}

Let $D$ denote an {\em event space\/}.
More precisely an element of $D$ is a triple whose first element is a
Profile IDentifier (PID) chosen from a finite set $\calI$, whose second
element is a label chosen from a finite set of labels $\calL$,
and whose third element is an element of
$S$, a set of events associated with PIDs.
In short, $D = \calI\times\calL\times S$, where $\calI$ is the set of
PIDs and $\calL$ is the set of labels.

We use heavily the facts that $\calI$ and $\calL$ are finite sets.

We assume that $S$ is a  semigroup with unit~$1$ generated by
$S_0\subseteq S$.
For example, $S_0$ might be a set of transactions and $S$ might be
sequences of transactions.
Multiplication in $S$ might be concatenation, or some operation related
to the structure of the data represented by $S$.

A {\em labeled learning set\/} is an element of $\w{D}$, the
set of words $w=d_1\cdots d_k$ of events in $D$.

A {\em labeled learning sequence\/} is a sequence $\{w_1,w_2,
\ldots\}$ of labeled learning sets;
a corresponding {\em formal labeled learning series\/} is a formal
series
\[
\sum_{w} p_{w}w.
\]

Let $H=k\w{D}$ denote the vector space with basis $\w{D}$, and
$kS$ denote the vector space with basis $S$.
Then $H$ is an algebra whose multiplication is induced by the semigroup
structure of $\w{D}$, which is simply concatenation,
and $U=kS$ is an algebra whose structure is induced by the semigroup
structure of $S$.

Let $\oH$ denote the space of formal labeled learning series.
For $(i,\ell)\in\calI\times\calL$ define the map
$\pi_{(i,\ell)}:\oH\longrightarrow U^{*}$ as follows:
first, define $\pi_{(i,\ell)}(p)(s)=p((i,\ell,s))$ for $p\in D$ and $s\in S$;
then, extend $\pi_{(i,\ell)}$ to $\w{D}$ multiplicatively;

We have that $U=kS$ is a bialgebra, with coproduct given by
$\Delta(s) = 1\otimes s + s\otimes 1$ for $s\in S_0$, and with
augmentation $\epsilon$ defined by $\epsilon(1)=1$, $\epsilon(s)=0$ for all
non-identity elements $s\in S$.
We will view $S$ as acting on a state space.
Since $U$ is primitively generated, $U\cong U(P(U))$ (recall that
$P(U)=\{\,x\in U \mid \Delta(x)=1\otimes x+x\otimes1\,\}$
is a Lie algebra, and that $U(L)$ is the universal enveloping algebra of
the Lie algebra $L$ \cite{jacobson}).
We put a bialgebra structure on $H$ by letting
$\Delta((i,\ell,s)) =
\sum_{(s)}(i,\ell,s_{(1)})\otimes(i,\ell,s_{(2)})$
where $\Delta(s)=\sum_{(s)}s_{(1)}\otimes s_{(2)}$,
and $\epsilon((i,\ell,s))=\epsilon(s)$, for
$i\in\calI$, $\ell\in\calL$, $s\in S$, and extending multiplicatively to $\w{D}$

A {\em simple formal learning series\/} is an element $p\in U^{*}$.
We can think of a simple learning series $p$ as an infinite series
$\sum_{s\in S}c_s s$.
Essentially, a simple formal learning series is a formal labeled
learning series, but without the labels and PIDs.

\section{Construction of the state space}\label{stspthmsect}

We are concerned whether $p\in U^{*}$, or some finite set $\{p_\alpha\}\subset
U^{*}$, arises from a finite dimensional state space $X$.
The reason we work with a finite set of elements of $U^{*}$ rather
than with a single one is that this allows us to deal with individual
profiles that get aggregated into the full dataset.

Since $U$ is primitively generated, we know that $U\cong U(P(U))$.

\begin{rmk}
{\rm
If $H$ is any bialgebra, we have a left $H$-module action of $H$ on
$H^{*}$ defined by $h\lact p(k) = p(kh)$ for $p\in H^{*}$, $h$, $k\in
H$, and a right $H$-module action of $H$ on
$H^{*}$ defined by $p\ract h(k) = p(hk)$ for $p\in H^{*}$, $h$, $k\in
H$.
}
\end{rmk}

The following definition is from \cite{GL:1992}.
\begin{dfn}
We say that the simple formal learning series $p\in U^{*}$ has {\em finite Lie
rank\/} if $\dim P(U)\lact p$ is finite.

Let $R$ be a commutative algebra with augmentation $\epsilon$, and let
$f\in R$.
We say that $p\in U^{*}$ is {\em differentially produced by
the pair $(R,f)$\/} if
\begin{enumerate}
\item there is right $U$-module algebra structure $\cdot$ on $R$;
\item $p(u) = \epsilon(f\cdot u)$ for $u\in U$.
\end{enumerate}
\end{dfn}

A basic theorem on the existence of a state space is the following,
which is a generalization of Theorem 1.1 in \cite{GL:1992}.
In this theorem, the state space is a vector space with basis
$\{x_1,\ldots,x_n\}$.

\begin{theorem}\label{statetheorem}
Let $p_1$, \ldots, $p_r\in U^{*}$.
Then the following are equivalent:
\begin{enumerate}
\item\label{ThmItem1} $p_k$ has finite Lie rank
for $k=1$, \ldots, $r$;
\item\label{ThmItem2} there is an augmented algebra $R$
for which $\dim\,(\Ker\epsilon)/(\Ker\epsilon)^2$ is finite,
and for all $k$, there is $f_k\in R$ such that $p_k$ is
differentially produced by the pair $(R,f_k)$;
\item\label{ThmItem3} there is a subalgebra $R$ of $U^{*}$ which
is isomorphic to $k[[x_1,\ldots,x_n]]$, the algebra of formal power
series in $n$ variables,
and for all $k$, there is $f_k\in R$ such that $p_k$ is
differentially produced by the pair $(R,f_k)$.
\end{enumerate}
\end{theorem}

\pf
We first prove that part~(\ref{ThmItem1}) of
Theorem~\ref{statetheorem} implies part~(\ref{ThmItem3}).
Given $p_1$, \ldots, $p_r\in U^{*}$, we define three basic
objects:
\begin{eqnarray*}
L &=& \{\, u \in P(U) \mid u\lact p_k = 0, \mbox{ for all }k,\}
\\ J &=& UL \\
\jperp &=& \{\,q\in U^{*} \mid q(j)=0 \mbox{ for all } j \in J \}.
\end{eqnarray*}
Since $L\subseteq P(U)$, it follows that $J$ is a coideal,
that is, that $\Delta(J)\subseteq J\otimes U + U\otimes J$.
Therefore $\jperp\cong(U/J)^{*}$ is a subalgebra of $U^{*}$.
We will show that $\jperp$ is
isomorphic to a formal power series algebra.

\begin{lemma}\label{lemma-a}
If
$\dim \sum_k P(U)\lact p_k = n$,
then $\jperp$ is a subalgebra of $U^{*}$ satisfying
\[
\jperp \cong k[[x_1,\ldots, x_n]].
\]
\end{lemma}

\pf
Note that $L$ is the kernel of the map
\[
P(U)\longrightarrow\bigoplus_k P(U)\lact p_k,\qquad
u\mapsto \bigoplus_k u\lact p_k.
\]
and $L$ has finite codimension $n$.
Choose a basis $\{ e_1, e_2, \ldots \}$ of $P(U)$ such
that $\{ e_{n+1}, e_{n+2}, \ldots \}$ is a basis of $L$.
Note that if $\ebar_i$ is
the image of $e_i$
under the quotient map
$P(U) \rightarrow P(U)/L$, then
$\{ \ebar_1$, \ldots, $\ebar_n \}$
is a basis for $P(U)/L$.

By the Poincar\'e-Birkhoff-Witt Theorem, $U$ has
a basis of the form
\[
\{\, e_{i_1}^{\alpha_{i_1}} \cdots e_{i_k}^{\alpha_{i_k}} \mid
i_1 < \cdots < i_k\mbox{ and } 0<\alpha_{i_r}\,\}.
\]
Since the basis $\{e_i\}$ of
$P(U)$ has been chosen so that $e_i\in L$ for $i>n$, it follows that
the monomials
$\{\,e_1^{\alpha_1}\cdots e_n^{\alpha_n}\mid \alpha_k\ge0\,\}$ are a basis for a
vector space complement to $J$.
It follows that
\[
\{\,\ebar_1^{\alpha_1}\cdots\ebar_n^{\alpha_n} \mid
\alpha_1,\ldots,\alpha_n\ge0\,\}
\]
is a basis for $U/J$.
It now follows that the elements
\[
x_{\alpha} =
\frac{x^{\alpha}}{\alpha !} =
\frac{ x_{1}^{\alpha_{1}} \cdots x_{n}^{\alpha_{n}}}
{\alpha_{1}! \cdots \alpha_{n}! }
\]
are in $\jperp\subseteq U^{*}$,
where $x_i\in U^{*}$ is defined by
\[
x_i(e_{i_1}^{\alpha_{i_1}} \cdots e_{i_k}^{\alpha_{i_k}}) =
\left\{
\begin{array}{rl}
1 & \mbox{if $e_{i_1}^{\alpha_{i_1}} \cdots e_{i_k}^{\alpha_{i_k}}=e_i$,}\\
0 & \mbox{otherwise.}
\end{array}
\right.
\]
The subalgebra $\jperp$ consists
precisely of the closure in $U^{*}$ of the span of these
elements. In other words,
\[
\jperp \cong k[[x_1,\ldots, x_n]],
\]
completing the proof.

\medskip

We will use the following facts from the proof of
Lemma~\ref{lemma-a}:
suppose that $\{e_1$, \ldots, $e_n$, $\ldots\}$ is a basis for $P(U)$
such that
$\{e_{n+1}$, \ldots $\}$ is a basis for $L$.
Let $\{e^{\alpha}\}$ be the corresponding Poincar\'e-Birkhoff-Witt
basis.
Denote $\jperp$ by $R$.
Then $R\cong k[[x_1$, \ldots, $x_n]]$, and
$x_1^{\alpha_1}\cdots x_n^{\alpha_n}/{\alpha_1}!\cdots{\alpha_n}!$
is the element of the dual (topological) basis of $U^{*}$
to the Poincar\'e-Birkhoff-Witt basis $\{e^{\alpha}\}$ of $U$,
corresponding
to the basis element $e_1^{\alpha_1}\cdots e_n^{\alpha_n}$.

We now collect some properties of the ring of formal power
series $R$ which will be necessary for the proof of
Theorem~\ref{statetheorem}.

\begin{lemma}\label{lemma-b}
Assume $p\in U^{*}$ has finite Lie rank, and
let $R\subseteq U^{*}$, $e^{\alpha}\in U$, and $x^{\alpha}\in R$ be
as in Lemma~\ref{lemma-a}.
Define
\[
f = \sum_{\alpha=(\alpha_1,\ldots,\alpha_n)}
c_{\alpha}x^{\alpha} \in R,
\]
where $c_{\alpha} = \frac{\textstyle p(e^{\alpha})}{\textstyle\alpha!}$.
Then
\begin{enumerate}
\item\label{AssMeas} $U$ measures $R$ to itself via $\ract$;
\item\label{AssReal} $p(u)=\epsilon(f\ract u)$ for all $u\in U$.
\end{enumerate}
\end{lemma}

\pf
We begin with the proof of part~(\ref{AssMeas}).
Since $U$ measures $U^{*}$ to itself and $R\subseteq U^{*}$,
we need show only that $R\ract U\subseteq R$.
Take $r\in R$, $u\in U$ and $j\in J$.
We have $ (r\ract u)(j) = r(uj)$.
Since $J$ is a left ideal, $uj\in J$, so $r(uj)=0$, so
$r\ract u \in\jperp = R$.
This proves part~(\ref{AssMeas}).

We now prove part~(\ref{AssReal}).
Let $e^{\alpha}=e_{i_1}^{\alpha_{i_1}} \cdots e_{i_k}^{\alpha_{i_k}}$
be a Poincar\'e-Birkhoff-Witt basis element of $U$.
Since $e^{\alpha}\in J$
unless $\{i_1,\ldots,i_k\}\subseteq\{1,\ldots,n\}$,
$p(e^{\alpha})=0$
unless $\{i_1,\ldots,i_k\}\subseteq\{1,\ldots,n\}$.
Also $\epsilon(f\ract e^{\alpha})=f\ract e^{\alpha}(1)=
f(e^{\alpha}1)=f(e^{\alpha})=0$
unless $\{i_1,\ldots,i_k\}\subseteq\{1,\ldots,n\}$.
Now suppose $\{i_1$, \ldots, $i_k\}\subseteq\{1,\ldots,n\}$.
We have in this case that
$p(e^{\alpha})=\alpha!c_{\alpha}=f(e^{\alpha})=
f\ract e^{\alpha}(1)=\epsilon(f\ract e^{\alpha})$.
Since $\{e^{\alpha}\}$ is a basis for $U$, this completes the proof
of part~(\ref{AssReal}) of the lemma.

\medskip

\begin{corollary}
Under the assumptions of Lemma~\ref{lemma-b}, $f=p$.
\end{corollary}

Lemmas~\ref{lemma-a} and~\ref{lemma-b} yield that
part~(\ref{ThmItem1}) implies part~(\ref{ThmItem3}) in
Theorem~\ref{statetheorem}.
It is immediate that
part~(\ref{ThmItem3}) implies part~(\ref{ThmItem2}).

We now complete the proof of Theorem~\ref{statetheorem} by proving that
part~(\ref{ThmItem2}) implies part~(\ref{ThmItem1}).

Let $x_1,\ldots,x_n\in\Ker\epsilon$ be chosen so that
$\{\xbar_1,\ldots,\xbar_n\}$ is a basis for
$(\Ker\epsilon)/(\Ker\epsilon)^2$.
If $f\in R$ and $u\in U$, then
\[
f\cdot u=q_0(u)1+\sum_{i=1}^{n}q_i(u)x_i+g(u),
\]
where $q_i\in U^{*}$ and $g(u)\in (\Ker\epsilon)^2$.
Let $\ell\in P(U)$.
Since $U$ measures $R$ to itself and $\Delta(\ell)=1\otimes\ell+\ell\otimes1$,
the map $f\mapsto f\cdot\ell$ is a derivation of $R$.

Now let $f_k\in R$ be the element such that
\[
p_k(u)=\epsilon(f_k\cdot u).
\]
Then
\begin{eqnarray*}
f_k\cdot ul & = & (f_k\cdot u)\cdot\ell \\
          & = & q_{k,0}(u)1\cdot l+\sum_{j=1}^n
                q_{k,j}(u)x_j\cdot l + g_k(u)\cdot l.
\end{eqnarray*}
Since the map $f\mapsto f\cdot\ell$ is a derivation,
$1\cdot\ell=0$, and
since $g_\alpha(u)\in (\Ker\epsilon)^2$, $g_\alpha(u)\cdot
\ell\in\Ker\epsilon$. It follows that
\begin{eqnarray*}
\ell\lact p_k(u) & = & p_k(u\ell)\\
            & = & \epsilon(f_k\cdot u\ell)\\
            & = & \sum_{j=1}^nq_{k,j}(u)\epsilon(x_j\cdot\ell).
\end{eqnarray*}
Therefore $P(U)\lact p_k\subseteq \sum_{j=1}^nkq_j$, so $p_k$
has finite Lie rank.
This completes the proof of Theorem~\ref{statetheorem}

\medskip

\begin{dfn}
A series $p\in H$ for which the set
\[
\{\,p_{(i,\ell)} = \pi_{(i,\ell)}(p) \mid \ell\in\calL, \,\,i\in\calI\,\}
\]
satisfies
the conditions of Theorem~\ref{statetheorem} is called {\em regular\/}.
\end{dfn}

We have shown how to construct a state space $X$ and a right
$U$-module algebra $R$ of observations for a regular series.

Although the $R$ we have constructed is a power series algebra, for
applications we will often use some other right $U$-module algebra of
functions on $X$.
We will assume that we have an action of $S$ on $X$ which induces
the action of $U$ on $R$, that is, that $R$ is a $U$-module algebra.

\section{Learning sets of profiles and realizations}

Let $\calI$ be a finite set of PIDs, and let $\calL$ be a finite set of
labels.
Let $H$ and $U$ be the bialgebras described in Section~\ref{intro},
$X$ be the corresponding state space as described in
Section~\ref{stspthmsect},
and $R$ be a right $U$-module algebra of functions from $X$ to $k$.

\begin{dfn}
A {\em classifier\/} is a function $f:X\longrightarrow\calL$.
A {\em learning set of profiles\/} is a function
$\chi:\calI\longrightarrow\calL\times X$, that is, a finite set
$\{(\ell_j,x_j)\}$, where $\ell_j\in\calL$, $x_j\in X$, and $j\in\calI$.
\end{dfn}

Note that a classifier is a model as defined in Section 1.
We denote the set of classifiers by $\calF$ and the set of learning sets of
profiles by $\calC$.
$\w{D}$ acts on $\calC$ as follows.
If $d = (i,l,s) \in D$ and $\chi = \{(l_j,x_j)\}$,
define
$\chi\cdot d = \{(\ell_j,x_j)\cdot d\}$, where
\[
(\ell_j,x_j)\cdot d = \left\{
\begin{array}{ll}
(\ell,x_j\cdot s) &\mbox{if $i=j$} \\
(l_j,x_j)      &\mbox{otherwise.}
\end{array}
\right.
\]
That is, the event $d = (i,\ell,s)$ acts on the learning set of profiles
$\chi=\{(l_j,x_j)\}$
by acting on the individual points $(\ell_j,x_j)$ as follows:
if $j\not=i$ the point is unchanged; if $j=i$ the point $x_j$ is moved
to $x_j\cdot s$ and the label is changed to $\ell$.

A {\em pairing\/} $\lppt f,\chi\rppt$ between classifiers and learning sets of
profiles can be given as follows.
Let $f:X\longrightarrow\calL$ be a classifier, and $\chi=\{(\ell_i,x_i)\}$
be a learning set of profiles.
Then
\begin{equation}\label{profdef}
\lppt f,\chi\rppt = \frac
{|\{\,i\in\calI\mid f(x_i)=\ell_i\,\}|}{|\calI|}.
\end{equation}
Note that $0\le \lppt f,\chi\rppt\le 1$.
This pairing is a measure of how well the classifier $f$ predicts
the actual data represented by $\chi$.

We define the notion of {\em realization\/} as follows.

\begin{dfn}\label{realizationdef}
Let
\[
f:X\longrightarrow \calL
\]
be a classifier, let
\[
\chi: \calI \longrightarrow \calL \times X
\]
be a learning set of profiles, and let $\lppt-,-\rppt$ be a pairing.
We say that the triple $(X, f, \chi)$ is a {\em realization\/}
of the series $p\in\oH$ if
\[
p_h = \lppt f, \chi\cdot h\rppt.
\]
\end{dfn}

Note that the classifier $\lppt f,\chi\cdot h\rppt$ defined in Equation~(\ref{profdef})
is bounded, in fact
\[
0\le\lppt f,\chi\cdot h\rppt\le1.
\]
Recall that $p = \sum_{h\in \w{D}} p_h h$ is the formal series of which
we are studying realizations.

\begin{lemma}\label{approxlemma}
Fix a finite learning set $\chi$, and fix
$A\subseteq {\bbR}^n$.
Suppose that there is a map $M:A\longrightarrow \calF$ such that
$\lppt M(a),\chi\cdot h\rppt$
is a bounded function of $a\in A$,
and $p\in\oH$ for which there is a
state space $X$ and a ring of functions $R$ as described in
section~\ref{stspthmsect}.
Assume that $p_h$, $h\in\w{D}$, is bounded.
Let
\[
\widetilde{M}(a) =  \sup_{h\in \w{D}}|p_h-\lppt M(a),\chi\cdot h\rppt|,
\]
Then for all $\epsilon>0$ there exists $a_0\in A$ such that
$|\widetilde{M}(a_0) - \inf_{a\in A}\widetilde{M}(a)|<\epsilon$.
\end{lemma}
Note that the hypothesis on $M$ includes models which are polynomials,
tree classifiers, neural nets, and splines.

\pf

Since everything in its definition is bounded, $\widetilde{M}(a)$ exists
and is bounded.
If $P:A\longrightarrow{\bbR}$ is any bounded function, then there is
$a_0\in
A$ such that $P(a_0)$ is within $\epsilon$ of $\inf_{a\in A}P(a)$.

\medskip

Note that for any realization $M(a)$ of $p$, we have that
\[
\widetilde{M}(a) = \sup_{h\in\w{D}}|p_h - \lppt M(a),\chi\cdot h\rppt|
\]
measures how well $M(a)$ realizes $p$.
so that
$\inf_{a\in A}\widetilde{M}(a)$
is the lower bound for the ``goodness'' of any realization.
The lemma says that this lower bound can be approximated arbitrarily closely.

\begin{theorem}\label{realthm}
Let $p:k\w{D}\longrightarrow k$ be such that $p_h$ is bounded, and let
$M:A\longrightarrow\calF$ be a parametrized classifier such that $\lppt
M(a),\chi\rppt$ is a bounded function of $a$.
Then for all $\epsilon>0$ there is a realization $p_0=M(a_0)$ of $p$
such that the ``goodness'' of the realization afforded by $p_0$ is
within $\epsilon$ of the lower bound, that is,
$|\widetilde{M}(a_0) - \inf_{a\in A}\widetilde{M}(a)|<\epsilon$.
\end{theorem}

\pf

Theorem~\ref{realthm} follows immediately from
Corollary~\ref{approxlemma}.

\medskip

\section{Parametrized realizations}

In this section we consider an event space $D$, a realizable labeled
learning
series $p$, a state space $X$, and an algebra of functions $R$ from the
state space $X$ to $k$.

Denote by $\calC$ learning sets of profiles
and denote by $\calF$ the set of functions from $X$ to the finite set
of labels $\calL$.
Fix a vector space of parameters $A$, and a map
\[
M:A\longrightarrow\calF
\]
giving a parametrized family of models.

In this section we study parametrized realizations of
formal series $p\in \oH$ of learning sets.

Compare Definition~\ref{parmrealdef} to Definition~\ref{realizationdef}
in which realizations are defined.

\begin{dfn}\label{parmrealdef}
A {\em parametrized realization\/} of a bounded function $p\in \oH$
is:
\begin{enumerate}
\item A vector space of parameters $A$.
\item A parametrized family of models $M:A\longrightarrow\calF$.
\end{enumerate}
If $A$ is a finite dimensional vector space, we say that the realization
is {\em$A$-finite\/}.
\end{dfn}

\medbreak
Theorem~\ref{realizationtheorem} below gives a
finiteness condition on the action of $A$ on $p\in \oH$ which
gives an $A$-finite realization.

For $f=M(a)\in\calS$ and $\ell\in\calL$, let $f_\ell$ be defined by
\[
f_\ell(x)=\left\{
\begin{array}{rl}
l       & \mbox{if $f(x)=\ell$,}\\
\star   & \mbox{otherwise,}
\end{array}\right.
\]
where $\star$ is unequal to any label $\ell\in\calL$.
Let $p_\ell(h)$ be defined by
\[
p_\ell(h)=\lppt f_\ell,\chi\cdot h\rppt.
\]

\begin{theorem}\label{realizationtheorem}
Let $p\in H^{*}$ be a formal sum of learning sets
and $M:A\longrightarrow \calF$
a family of labeled models parametrized by $A$.
Assume:
\begin{enumerate}
\item\label{Thm2item1} there exists $f\in\Im M$ such that
\[
p(h) = \lppt f, \chi\cdot h\rppt,
\]
\item\label{Thm2item2}  $\{\,\beta\in\Adual \mid
\beta\lact p_\ell=0\,\}$ is a
subspace of $\Adual$ of finite codimension which is closed in the
compact open topology for all $\ell\in\calL$.
\end{enumerate}
\par\noindent
Then there exists an $A$-finite realization of $p$.
\end{theorem}

Note that Theorem~\ref{realthm} gives the existence of a
realization which approximates the desired one.

\pf
We define three basic objects:
\begin{eqnarray*}
L_\ell   & = & \{\,\beta\in A^{*} \mid \beta\lact p_\ell = 0 \,\} \\
J_\ell   & = & k[\Adual]L_\ell \\
\jperp_\ell & = & \{\,q\in k[\Adual]^{*} \mid q(j)=0 \mbox{ for all } j \in
J_\ell\,\}.
\end{eqnarray*}
We have that $J_\ell$ is a coideal in the Hopf algebra $k[\Adual]$
generated by primitive elements in $\Adual$, that is,
that $\Delta(J_\ell)\subseteq J_\ell\otimes k[\Adual] + k[\Adual]\otimes J_\ell$.
Therefore $\jperp_\ell\cong(k[\Adual]/J_\ell)^{*}$ is a subalgebra of
$k[\Adual]^{*}$. We will show that $\jperp_\ell$ is
isomorphic to a formal power series algebra in finitely many variables.

From hypothesis~(\ref{Thm2item2}) we have that
 $L_\ell^\perp = (\Adual/L_\ell)^{*}$ is finite dimensional
subspace of $A$.

\begin{lemma}\label{lemma-a2}
If $\dim L_\ell^\perp=n_\ell$,
then $\jperp_l$ is a subalgebra of $k[\Adual]^{*}$ satisfying
\[
\jperp_l \cong k[[a_1,\ldots, a_{n_\ell}]],
\]
where $\{a_1,\ldots,a_{n_\ell}\}$ is a basis for $L_\ell^\perp$.
\end{lemma}

\pf
The subspace $L_\ell$ is a closed subspace of $\Adual$ of finite
codimension $n$, so that $(\Adual/L_\ell)^{*}=L_\ell^\perp$ is a finite
dimensional subspace of $A$. Let $(\Adual/L_\ell)^{*}$ have basis
$\{a_1,\ldots,a_{n_\ell}\}$.
Choose $\beta_{a_i}\in A^{*}$ with $\beta_{a_i}(a_j)=\delta_{ij}$.
Now choose a basis $\calB$ of $\Adual$ such that
$\calB\supseteq\{\beta_{a_1}$, \dots $\beta_{a_n}\}$ and
$\calB' =
\calB\setminus\{\beta_{a_1},\beta_{a_2},\ldots,\beta_{a_{n_\ell}}\}$
is a basis of $L_\ell$.
We have that $k[\Adual]$ has a basis
\[
\{\,\beta_{i_1}^{\alpha_{i_1}} \cdots \beta_{i_k}^{\alpha_{i_k}} \mid
\beta_k\in\calB, \mbox{ }i_1<\cdots< i_k,\mbox{ and }
0<\alpha_{i_r}\,\}.
\]
By the choice of the basis of $\Adual$,
$J_\ell$ will have a basis of the form
\[
\beta_{i_1}^{\alpha_{i_1}} \cdots \beta_{i_k}^{\alpha_{i_k}}
\]
with at least one $\beta_k\in\calB'$.
It follows that
\[
\{\,\beta_{a_1}^{\alpha_1}\cdots\beta_{a_{n_\ell}}^{\alpha_{n_\ell}} \mid
\alpha_1,\ldots,\alpha_{n_\ell}\ge0\,\},
\]
where we denote by $\beta_{a_k}$ the image of that element in
$k[\Adual]/J_\ell$, is a basis for $k[\Adual]/J_\ell$.
It now follows that elements of the form
\[
a^{\alpha} =
 a_{i_1}^{\alpha_{i_1}} \cdots a_{i_{n_\ell}}^{\alpha_{i_{n_\ell}}}
\]
are in $\jperp_l\subseteq U^{*}$.
and that $\jperp_l$ consists
precisely of the closure in $k[\Adual]^{*}$ of the span of such
elements.
In other words,
\[
\jperp_\ell \cong k[[a_1,\ldots, a_{n_\ell}]],
\]
completing the proof.

\medskip

By Lemma~\ref{lemma-a2} each $p_\ell$ depends on a finite dimensional space
of parameters $A_{(\ell)}$.
Let $A^0$ be the finite dimensional subspace which is the spanned
by the union of these finite dimensional subspaces.
Since $p(h)=\sum_{\ell\in\calL} p_\ell(h)$, $p(h)$ depends only on
parameters in $A^0$.

Now $\lppt f,\chi\cdot h\rppt$ depends only on parameters in $A^0$.
We
may choose the other parameters which are linearly independent from
$A^0$ arbitrarily.
In other words we may choose $f_0$ so that it depends only on
the parameters in $A^0$.

This completes the proof of Theorem~\ref{realizationtheorem}.

\end{document}